%% file: root.tex
\documentclass[a4paper, 10 pt, conference]{ieeeconf} 

\IEEEoverridecommandlockouts
\usepackage{tikz}
\usetikzlibrary{shapes}
\usetikzlibrary{arrows}

\usepackage[style=ieee,mincitenames=1,isbn=false]{biblatex}
\AtEveryBibitem{
\ifentrytype{inproceedings}{
 	\clearlist{address}
 	\clearlist{publisher}
 	\clearname{editor}
 	\clearlist{organization}
 	\clearfield{url}  
 	\clearfield{doi}  
 	\clearfield{pages}  
 	\clearlist{location}
 }{}
 }
 
\usepackage{bbm}
\usepackage{amsfonts}
\usepackage[makeroom]{cancel}
\usepackage{amsmath}
\newtheorem{proposition}{Proposition}
\usepackage{booktabs}
\usepackage{bm}
\usepackage{siunitx}
\usepackage{dsfont}

\usepackage{subcaption}
\usepackage{graphicx}

\usepackage{cleveref}

\usepackage{algorithm}
\usepackage{algorithmicx}
\usepackage[noend]{algpseudocode}

\usepackage[keeplastbox]{flushend}

\title{\LARGE \bf Scalable Autonomous Vehicle Safety Validation through \\ Dynamic Programming and Scene Decomposition }

\author{Anthony Corso,$^{1}$ Ritchie Lee,$^{2}$ and Mykel J. Kochenderfer$^{1}$%
\thanks{A. Corso and M. J. Kochenderfer are with the Aeronautics and Astronautics Department, Stanford University. e-mail: \{acorso,mykel\}@stanford.edu}%
\thanks{R. Lee is with the Robust Software Engineering group at NASA Ames Research Center. e-mail: ritchie.lee@nasa.gov}%
}

\begin{document}
\maketitle
\thispagestyle{empty}
\pagestyle{empty}

\begin{abstract}

An open question in autonomous driving is how best to use simulation to validate the safety of autonomous vehicles. Existing techniques rely on simulated rollouts, which can be inefficient for finding rare failure events, while other techniques are designed to only discover a single failure. In this work, we present a new safety validation approach that attempts to estimate the distribution over failures of an autonomous policy using approximate dynamic programming. Knowledge of this distribution allows for the efficient discovery of many failure examples. To address the problem of scalability, we decompose complex driving scenarios into subproblems consisting of only the ego vehicle and one other vehicle. These subproblems can be solved with approximate dynamic programming and their solutions are recombined to approximate the solution to the full scenario.  We apply our approach to a simple two-vehicle scenario to demonstrate the technique as well as a more complex five-vehicle scenario to demonstrate scalability. In both experiments, we observed an increase in the number of failures discovered compared to baseline approaches.

\end{abstract}

\section{INTRODUCTION}

One common practice for automated vehicle (AV) safety validation is to maintain a suite of challenging driving scenarios that the vehicle must successfully navigate after each update to the driving policy. Although useful, this approach will miss any failures that are not already included in the test suite. Automated testing procedures that treat the vehicle as a black box must be developed to catch unknown and unexpected failure modes of the AV which could dramatically decrease testing time and improve the safety of autonomous vehicles. 

Much of the literature on black box testing focuses on falsification where inputs are generated that cause a system to violate a safety specification~\cite{Donze2010robust}. Those inputs serve as a counter example to the hypothesis that the system is safe. For autonomous driving, it is not feasible to create an agent that can avoid all possible accidents~\cite{shalev2017formal}, so rather than find any failure of an AV, it is preferable to find the most likely failures. Traditional falsification techniques do not consider the probability of the failures they find and are therefore ill-suited to this goal. Adaptive stress testing~\cite{lee2015adaptive} tries to find the most-likely failure of an autonomous system. This approach can improve the likelihood of discovered failures but does not necessarily explore the range of possible failures of the system. The goal of this work is to develop a safety validation approach that can reliably find all of the most relevant failures of an autonomous vehicle.

Our approach attempts to estimate the distribution over failures of an autonomous vehicle operating in a stochastic environment. If we assume that the vehicle's policy and simulator are Markov then we show that the problem simplifies to estimating the probability of failure at each state, a computation which can be performed using approximate dynamic programming (DP). Approximate DP is particularly effective at finding failures because it can start at a failure and work backward to see what led to it. Unfortunately, this approach has difficulty scaling to large state spaces. To improve  scalability, we use the structure of driving scenarios by decomposing the simulation into pairwise interactions between the ego vehicle and other agents on the road. These subproblems are tractable for approximate DP, and their solutions can be recombined to approximate the solution for the full problem. To account for the approximation error due to multi-agent interactions, we combine the subproblems using a learned set of weights.

We apply our approach to two driving scenarios: a simple two-vehicle scenario to demonstrate the effectiveness of DP, and a more complex five-vehicle scenario to demonstrate the favorable scaling of the approach. In both experiments, we observed increases in the number of failures discovered compared to baseline approaches, and the discovered failure had comparatively high likelihood. The main contributions of this work are:
\begin{itemize}
    \item A safety validation approach that estimates the distribution over failures using approximate DP.
    \item An algorithm for problem decomposition and reconstruction to scale approximate DP to complex driving scenarios.
    \item Demonstration of these techniques on two realistic driving scenarios and observation of a significant increase in rates of discovered failures. 
\end{itemize}

The remainder of the paper is organized as follows: \cref{sec:related_work} gives an overview of related work in the field of black-box validation for autonomous driving, \cref{sec:proposed_approach} describes our proposed technique in detail, \cref{sec:experiments} outlines the two experiments and describes our results, and \cref{sec:conclusions} concludes and discusses future work.

\section{Related Work}
\label{sec:related_work}
This section discusses safety validation of autonomous systems. We first give a brief overview of black-box falsification algorithms and then discuss approaches that were developed specifically for autonomous driving.

\subsection{Safety Validation of Black-Box Systems}
Falsification of black box systems involves finding inputs to the system that lead to violation of the system specifications. State-of-the-art approaches cast falsification as a global optimization problem over the input space~\cite{Donze2010robust} and try to solve it using surrogate models~\cite{Mathesen2019falsification}, deep reinforcement learning~\cite{Akazaki2018falsification}, genetic algorithms~\cite{zhao2003generating}, Monte Carlo tree search~\cite{zhang2018two}, or cross-entropy optimization~\cite{sankaranarayanan2012falsification}. Adaptive stress testing (AST)~\cite{lee2015adaptive,koren2018adaptive,corso2019adaptive,koren2019Efficient} frames the problem of falsification as a Markov decision process and uses reinforcement learning to find the most-likely failures of a system according to a prescribed probability model. The field of statistical model checking~\cite{agha2018survey} deals with estimating the probability of failure, and in doing so will find inputs to the system that cause it to fail.

Several approaches rely on sampling-based methods to discover failures. \Citeauthor{huang2019evaluation}~\cite{huang2019evaluation} use bootstrapping and importance sampling to obtain a low-variance estimate of the probability of failure. Another approach uses importance sampling via the cross-entropy method to increase the number of failures found in simulation~\cite{kim2016improving,okelly2018scalable}. \Citeauthor{uesato2018rigorous}~\cite{uesato2018rigorous} use previous versions of an autonomous agent to help find failures in the final version, an approach that works when when agents have learned behavior. Similar to the present work, \Citeauthor{Chryssanthacopoulos2010}~\cite{Chryssanthacopoulos2010} use DP to estimate the probability of failure. 

\subsection{Safety of Autonomous Vehicles}
Some work has focused on falsifying components of an autonomous vehicle such as Adaptive Cruise Control \cite{koschi2019computationally} or perception systems~\cite{cao2019adversarial,Balakrishnan2019specifying}. Other work has focused on the generation of critical test cases. For example \citeauthor{mullins2018adaptive}~\cite{mullins2018adaptive} identify regions of the input space that separate distinct types of autonomous agent behavior, and \citeauthor{althoff2018automatic}~\cite{althoff2018automatic} design adversarial agents to minimize the safe available driving space of the autonomous vehicle.

\section{PROPOSED APPROACH}
\label{sec:proposed_approach}
This section describes our approach to the safety validation problem. We start with the problem formulation and definition of notation. Then, we describe our technique for estimating the distribution over failures assuming we know the probability of failure from each state. Lastly, we describe how to compute that probability in a scalable way.

\subsection{Problem Formulation}

Suppose we wish to analyze the safety of a black-box autonomous system (system-under-test, or SUT) that operates in a stochastic simulated environment. The state of the SUT and the environment is $s \in \mathcal{S}$ and the disturbances $x \in \mathcal{X}$ are stochastic elements of the environment that influence the behavior of the SUT. A state-disturbance trajectory $\tau = \{ s_0, x_1, s_1 \ldots, x_{N}, s_{N}\}$ has a likelihood of occurrence $p(\tau)$. We define $E$ as the set of all failure states of the SUT and the notation $s_N \in E$ means that the trajectory $\tau$ ends in a failure. Let $T$ be the set of all terminal states where $E \subseteq T$.

We would like to know the distribution over failures
\begin{equation} 
f(\tau) = \frac{\mathds{1}\left\{ s_N \in E \right\} p(\tau)}{\mathbb{E}_p\left[ \mathds{1}\{ s_N \in E \} \right]}
\end{equation}
where $\mathds{1}$ is the indicator function and the denominator normalizes the distribution. Note that $f(\tau)$ is the minimum-variance importance sampling distribution for estimating the probability of failure.

\subsection{Estimating the Distribution Over Failures}

The space of all trajectories is exponential in the legnth of the trajectory, so it will be challenging to represent the distribution $f(\tau)$ directly. To reduce the dimensionality of the distribution we assume that the SUT and environment are Markov. The current disturbance $x$ and next state $s^\prime$ will only depend on the current state $s$ such that
\begin{equation}
    p(x, s^\prime \mid s) = \underbrace{p(x \mid s)}_{\rm disturbance \ model}\overbrace{p(s^\prime \mid s, x)}^{\rm dynamics}.
\end{equation}
If we also assume that the dynamics of the SUT and the environment are deterministic (i.e. all stochasticity is controlled through disturbances), then
\begin{equation}
    p(x, s^\prime \mid s) = p(x \mid s).
\end{equation}
With these assumptions, the distribution over failures only depends on $p(x \mid s)$ and is given by 
\begin{equation}
    f(\tau) = \frac{\mathds{1}\{s_N \in E\}}{\mathbb{E}_p\left[ \mathds{1}\{ s_N \in E \} \right]} \prod_{t=1}^N p(x_t \mid s_{t-1})\label{eq:ftau}.
\end{equation}

The Markov assumption allows us to find a distribution over disturbances, or stochastic policy, $\pi$ that generates sample trajectories (rollouts) distributed according to $f$. Let
\begin{equation}
    \pi(x \mid s) = \frac{p(x \mid s) v(s^\prime)}{\sum_{a'}  p(x' \mid s) v(s'')} = \frac{p(x \mid s) v(s^\prime)}{v(s)} \label{eq:opt_pol}
\end{equation}
where $v(s)$ is the probability of failure from state $s$ and $s''$ is the state reached from $s$ after applying disturbance $x'$. The second equality in \cref{eq:opt_pol} comes from the observation that the probability of failure in the current state is a sum of the probability of failure over possible next states, weighted by the likelihood of reaching that state. 

\begin{proposition}
Trajectories generated from rollouts of the policy $\pi$ will be distributed according to $f$.
\end{proposition}

\proof
Let $f^*(\tau)$ be the distribution induced by rollouts of the policy $\pi$. We will show that for any $\tau$, $f(\tau) = f^*(\tau)$. First, we define the Bellman equation that describes the probability of failure of a Markov system as
\begin{equation} v(s) = \left\{ \begin{array}{ll}
1 &  \text{if} \ s \in E \\
0 &  \text{if} \  s \notin E, \ s \in T \\
\sum_a p(x \mid s) v(s^\prime) &  \text{otherwise}
\end{array} \right. \label{eq:belman}\end{equation}
Then consider an arbitrary trajectory $\tau$ that has probability according to $f^*$ given by
\begin{equation}
    f^*(\tau) = \prod_{t=1}^{N} \pi(x_t \mid s_{t-1}) = \prod_{t=1}^{N} \frac{p(x_t \mid s_{t-1}) v(s_t)}{v(s_{t-1})} \label{eq:fstar}
\end{equation}
and probability according to $f$ given by \cref{eq:ftau}. There are two cases to consider.\\
\textbf{Case 1: } $\mathds{1}\{s_N \notin E\}$

Due to the indicator function in \cref{eq:ftau}, $f(\tau)=0$. With final state $s_N \in T$ and $s_N \notin E$, we have $v(s_N) = 0$. The last term in the product in \cref{eq:fstar} contains $v(s_N)$, making $f^*(\tau) = 0$.\\
\textbf{Case 2: } $\mathds{1}\{s_N \in E\}$ 

Considering the definition of $f^*(\tau)$, we have
\begin{align}
    f^*(\tau) &= \prod_{t=1}^{N} \frac{p(x_t \mid s_{t-1}) v(s_t)}{v(s_{t-1})} \\
    &= \frac{\cancelto{1}{v(s_N)}}{v(s_0)} \prod_{t=1}^{N} p(x_t \mid s_{t-1}) \\
    &= \frac{1}{\mathbb{E}_p\left[ \mathds{1}\{ s_N \in E \} \right]} \prod_{t=1}^{N} p(x_t \mid s_{t-1}) \\
    &= f(\tau)
\end{align}
where, in the second line, all of the $v(s_t)$ terms were canceled except $v(s_0)$ and $v(s_N)$, and \cref{eq:belman} was used to let $v(s_N) = 1$. In the third line, we observe that the probability of failure at the initial state $v(s_0)$ is equivalent to the expectation of failures $\mathbb{E}_p\left[ \mathds{1}\{ s_N \in E \} \right]$. Thus, in all cases, $f(\tau) = f^*(\tau)$.
\endproof

Assuming that the distribution over disturbances $p(x \mid s)$ is provided (either from domain knowledge or data), then the problem of computing the distribution over failures amounts to computing the probability of failure at each state $v(s)$. Additionally, once $v(s)$ is known, failures can be generated from any initial condition where a failure can be reached. 

\subsection{Computing the Probability of Failure}

The feasibility of computing the probability of failure $v(s)$ depends on the size of the state and disturbance spaces. If those spaces are discrete and relatively small, then DP can be used to compute $v$ to any desired level of accuracy. If the state space is continuous, but is small enough to be discretized, then local approximation DP can be used to estimate $v(s)$~\cite{kochenderfer2015decision}. As will be demonstrated by our experiments, this approach is feasible for interactions between two vehicles on the road. For more vehicles, discretizing the state space becomes intractable and we must rely on further approximation. 

When scaling to much larger state-spaces, we can leverage the structure of the problem to improve scalability. We propose to decompose a complicated driving scenario into pairwise interactions between the ego vehicle and other agents on the road, similar to the decomposition approach used by \citeauthor{bouton2019decomposition} \cite{bouton2019decomposition}. Each subproblem can then be solved for the probability of failure between the $i$th vehicle and the ego vehicle yielding $v_i(s^{(i)})$, where $s^{(i)}$ is the subset of the state representing only those vehicles. To combine the probability of failure from each of $m$ subproblems, we can use the transfer learning approach called attend, adapt and transfer (A2T)~\cite{rajendran2015attend}. A2T combines the solutions of $m$ problems with a solution learned from scratch $v_{\rm base}$ using a learned set of state-dependent attention weights $w(s)$. The estimated probability of failure for a state $s$, $\tilde{v}(s)$, is then given by 
\begin{equation}
\tilde{v}(s) = w_0(s) v_{\rm base}(s) + \sum_{i=1}^m w_i(s) v_i(s^{(i)}) \label{eq:A2T}
\end{equation}
where $w_i$ and $v_{\rm base}$ have parameters that can be learned.

\begin{figure}[!t]
\centering
\input{A2T_Network}
\caption{The A2T network. Dashed lines represent backpropagation for learning parameters. }
\label{fig:A2T_Network}
\vskip -0.5cm
\end{figure}
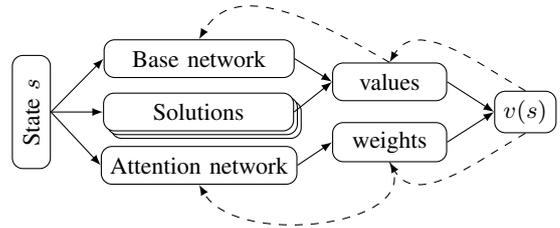

The use of attention weights allows A2T to learn which solutions are most relevant in which states. If none of the subproblems are providing a good estimate then the base network will learn a good estimate from scratch. The estimate from \cref{eq:A2T} can be represented as the network architecture shown in \cref{fig:A2T_Network}. The base network has two hidden layers each with \num{32} units and relu activations followed by a sigmoid activation to keep the output between \num{0} and \num{1}. The solutions take the state as input and give the probability of failure estimate for each subproblem. The attention network has one hidden layer with \num{32} units and a softmax layer to make sure the weights sum to 1. The base network output is concatenated to the subproblem solution outputs to create a vector of values that has $m+1$ components. The dot product is taken between the values and the $m+1$ weights to produce the final estimate of the probability of failure. 

\begin{algorithm}
\caption{MC evaluation with function approximation}
    \label{alg:mc_policy_eval}
\begin{algorithmic}[1]
    \footnotesize
    \Function{MCPolicyEval}{$\tilde{v}_\theta$, $N_{\rm iter}$, $N_{\rm samp}$, $\alpha$}
    \For{$N_{\rm iter}$ iterations}
        \State $S, G \gets$ Rollouts($\tilde{v}_\theta$, $N_{\rm samp}$) \label{line:rollouts}
        \State $J = \frac{1}{N_{\rm samp}} \sum_{j=1}^{N_{\rm samp}} (G_j - v(S_j))^2$ \label{line:mse}
        \State $\theta \gets \theta - \alpha \nabla_\theta J$ \label{line:update}
    \EndFor
    \State \textbf{return} $\tilde{v}_\theta$
    \EndFunction
\end{algorithmic}
\end{algorithm}

The network can be trained using rollouts from the full driving scenario to estimate the probability of failure. The training procedure we used is Monte Carlo policy evaluation with function approximation~\cite{sutton2018reinforcement} and is shown in \cref{alg:mc_policy_eval}. The algorithm takes as input the network that estimates the probability of failure $\tilde{v}_\theta$ with trainable parameters $\theta$, the number of training iterations $N_{\rm iter}$, the number of sampled transitions per iteration $N_{\rm samp}$, and the learning rate $\alpha$. On each iteration, a series of rollouts are performed (line \ref{line:rollouts}). The rollout policy is $\pi$ from \cref{eq:opt_pol} where $v(s)$ is replaced with the current estimate $\tilde{v}_\theta(s)$. As the estimate of the probability of failure is improved, the rollout policy will produce more failure examples. All of the states visited during the rollouts are concatenated into a vector $S$. The return is computed for each state $s_j \in S$ as
\begin{equation}
    G(s_j) = \mathds{1}\{s_N \in E \} \prod_{t=j}^N p(x_t \mid s_{t-1}) / \pi(x_t \mid s_{t-1}) \label{eq:return}
\end{equation} 
where $N$ is the length of the episode that contained state $s_j$. The estimate $G$ is a Bernoulli sample weighted by the likelihood ratio of the current sampling policy so the expected value of of $G(s)$ is the probability of failure from state $s$. The cost $J$ is the mean squared error between the estimated probability of failure $\tilde{v}_\theta(s)$ and $G(s)$ (line \ref{line:mse}). The parameters of the network are updated using the gradient of the cost function to improve the estimate (line \ref{line:update}).

\section{EXPERIMENTS}
\label{sec:experiments}

This section describes two experimental driving scenarios, a simple scenario with two vehicles, and a more complex scenario with five vehicles. The simulations were designed with AutomotiveSimulator.jl, an open-source julia package. Both simulations rely on the same road geometry and autonomous driving policy. The SUT is an autonomous vehicle referred to as the ego vehicle and a failure refers to any instance where the ego vehicle collides with another vehicle.

 The road geometry and initial vehicle configurations are pictured in \cref{fig:two_car_nominal,fig:two_car_collision,fig:five_car_nominal,fig:five_car_collision}. The driving scenario an unprotected left turn of the ego vehicle (in blue) onto a two-lane road. Other vehicles (referred to adversarial vehicles) are initialized on the through-road and can either continue straight or turn (the yellow dot represents a turn signal). The right-of-way rules are 1) vehicles on the through-road have right-of-way over vehicles turning on to the through-road, and 2) vehicles turning right have right-of-way over vehicles turning left.

The state of each vehicle can be described with four variables: position along the lane, velocity along the lane, a Boolean indicating if the turn signal is on, an integer indicating the lane. For approximate DP, the position and velocity were each discretized into \num{15} values and each vehicle can be in one of two lanes so each vehicle had a total of \num{900} states. 

 Each vehicle on the road including the ego vehicle, follows a modified version of the intelligent driver model (IDM)~\cite{Treiber2000congested}. The IDM is a vehicle-following algorithm that tries to drive at a specified velocity while avoiding collisions with leading vehicles. In our experiments, the IDM is parameterized by a desired velocity of \SI{29}{m/s}, a minimum spacing of \SI{5}{m}, a maximum acceleration of \SI{3}{m/s^2} and a comfortable braking deceleration of \SI{-2}{m/s^2}, and a simulation timestep of $\Delta t = \SI{0.18}{s}$. The IDM was modified with a rule-based algorithm (\cref{alg:intersection_navigation}) for navigating the T-intersection. Each vehicle reasons about right-of-way and turning intention of other vehicles based on the state of their blinker, and uses current vehicle speeds to calculate if the intersection is safe to cross.
 
 \begin{algorithm}
\caption{Intersection navigation algorithm}
    \label{alg:intersection_navigation}
\begin{algorithmic}[1]
    \footnotesize
    \Function{ComputeAcceleration}{$veh$, $scene$}
    \State $v \gets$ velocity($veh$)
    \State $v_{\rm lead}$, $\Delta s_{\rm lead} \gets$ leading\_vehicle($veh$, $scene$)
    \State $acc \gets$ IDM\_acceleration($\Delta s_{\rm lead}$, $v$, $v_{\rm lead}$) 
    \If{$veh$ does not have right of way} 
        \State $ttc$ $\gets$ time\_to\_cross\_intersection($veh$) 
        \State $\Delta s_{\rm int}$ $\gets$ distance\_to\_intersection($veh$)
        \If {$\Delta s_{\rm int} < \Delta s_{\rm lead}$} 
        \For{$agent$ in $scene$}
            \State $ttenter$ $\gets$ time\_to\_enter\_intersection($agent$) 
            \State $ttexit$ $\gets$ time\_to\_exit\_intersection($agent$) 
            \If{$ttenter < ttc$ and $ttexit + \epsilon > ttc$} 
                \State $acc \gets$ IDM\_acceleration($\Delta s_{\rm int}$, $v$, 0) 
                \State \textbf{break}
            \EndIf
        \EndFor
        \EndIf
    \EndIf
    \State \textbf{return} $acc$
    \EndFunction
\end{algorithmic}

\end{algorithm}

The disturbances in the environment correspond to disturbances to the deterministic actions of all adversarial vehicles. The disturbances and their corresponding probabilities are shown in \cref{tab:adversarial_disturbance_space}. The first produces no disturbance, so the adversary accelerates by $a_{\rm IDM}$, the acceleration computed by the modified IDM. The next four disturbances perturb the adversary's acceleration by an amount $\delta a \in [\SI{-3}{m/s^2}, \SI{3}{m/s^2}]$ so that the actual acceleration of the adversary is $a_{\rm IDM} + \delta a$. The next disturbance toggles the adversary's turn signal which is observed by other vehicles and used to determine the adversary's intention. The final disturbance changes the hidden adversary intention as to whether or not it will turn. 

The choice of a disturbance probability model should be driven by real-world driving data. In absence of that data, we chose a simple probability model that made disturbances rare according to their magnitude (see MC Probability in \cref{tab:adversarial_disturbance_space}). Medium slowdowns and speedups were give a probability of \num{1e-2} per timestep while major slowdowns and speedups, toggling the blinker, and toggling turn intention had a per-timestep probability of occurrence of \num{1e-3}. 

\begin{table}
    \centering
    \caption{Action space for adversarial vehicles}
    \label{tab:adversarial_disturbance_space}
    \begin{tabular}{@{}lrr@{}} 
        \toprule
        \textbf{Action} & \textbf{Acceleration} & \textbf{MC Probability} \\
        \midrule
        No disturbance &  \SI{0}{m/s^2} & \num{0.976}\\
        Medium slowdown & \SI{-1.5}{m/s^2} & \num{1e-2}\\
        Major slowdown & \SI{-3}{m/s^2} & \num{1e-3}\\
        Medium speedup & \SI{1.5}{m/s^2} & \num{1e-2}\\
        Major speedup & \SI{3}{m/s^2} & \num{1e-3}\\
        Toggle blinker & N/A & \num{1e-3 }\\
        Toggle turn intent & N/A & \num{1e-3} \\
        \bottomrule
    \end{tabular}
    \vskip -0.2in
\end{table}

The two metrics we chose to evaluate our approach are the rate of failures found and the log-likelihood of adversary disturbances for failure trajectories. The failure rates are computed from \num{1000} rollouts and the average log-likelihood of disturbances is computed from \num{100} failure examples. The mean and standard deviations are reported. We compare our approach against three baselines:
\begin{enumerate}
    \item Monte Carlo: rollouts with the true probability distribution over disturbances.
    \item Uniform importance sampling: rollouts with a uniform distribution over disturbances.
    \item Cross entropy method: rollouts with a distribution over disturbances that has been optimized using the cross entropy method~\cite{de2005tutorial}.
\end{enumerate}

\subsection{Two-Vehicle Interaction}
\begin{figure}
    \centering
   \begin{subfigure}[t]{0.5\columnwidth}
        \centering
        \includegraphics[width=0.98\textwidth, trim={10cm 20cm 20cm 0},clip]{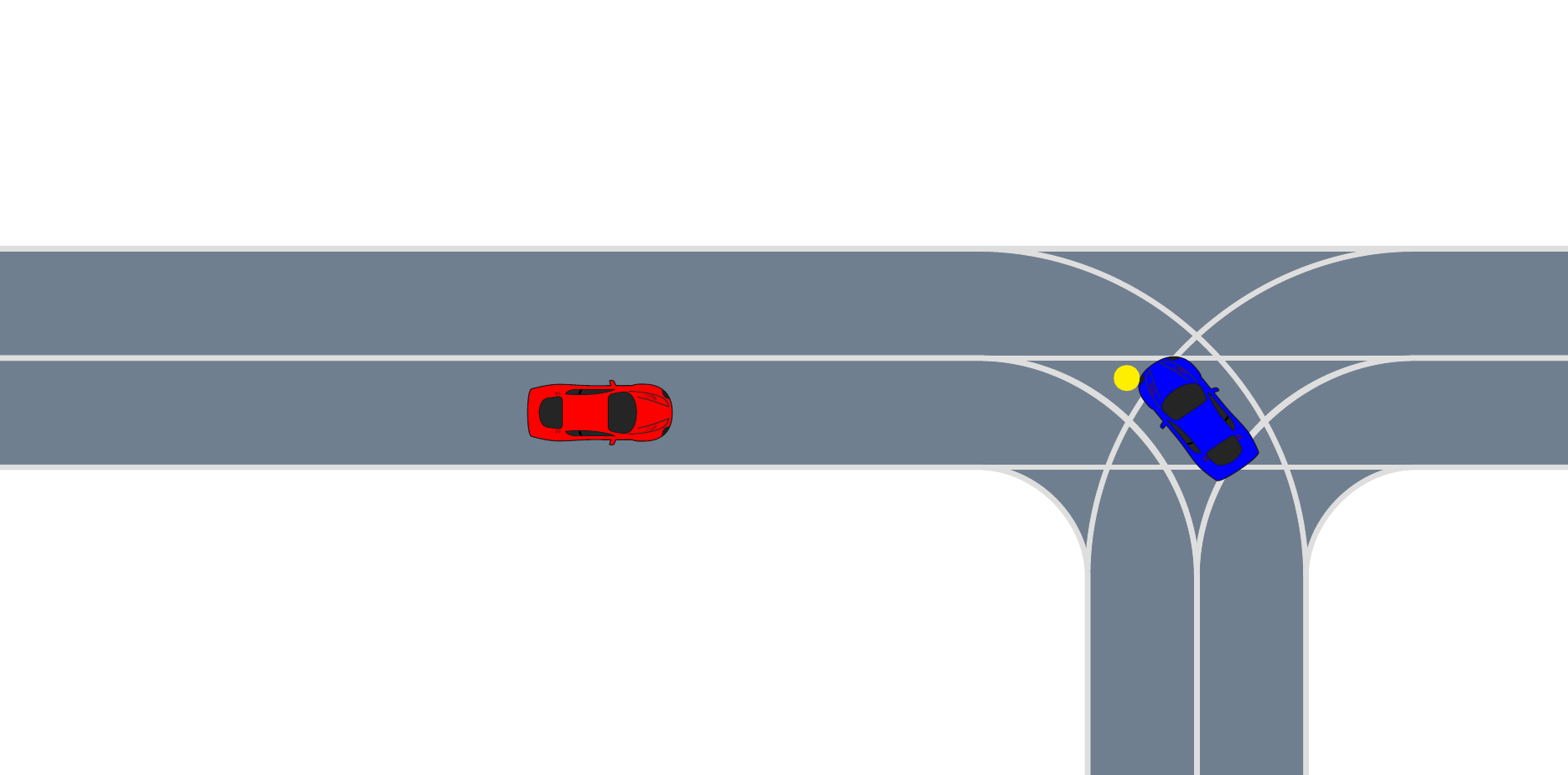}
    \end{subfigure}%
    \begin{subfigure}[t]{0.5\columnwidth}
        \centering
        \includegraphics[width=0.98\textwidth, trim={10cm 20cm 20cm 0},clip]{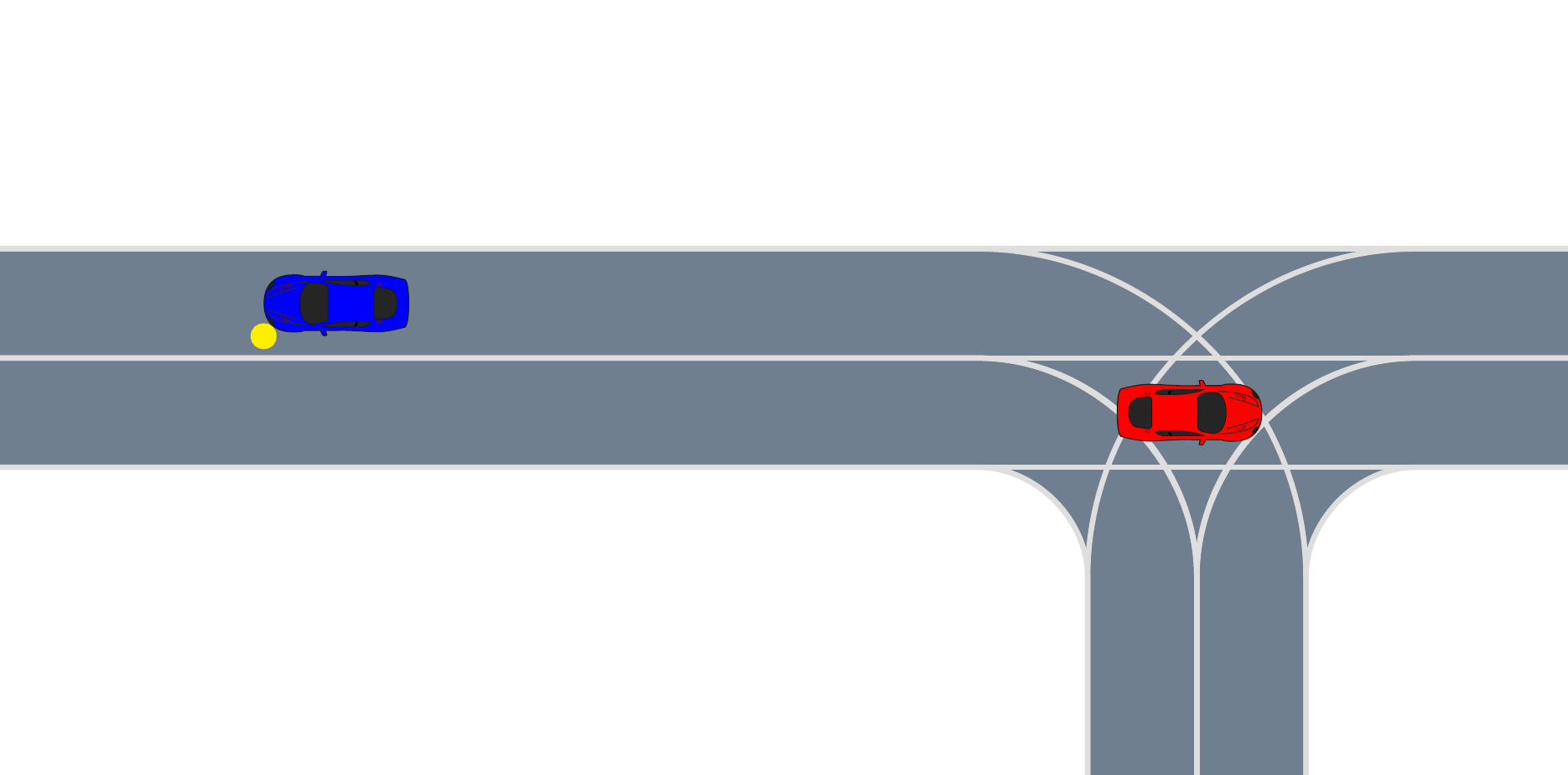}
    \end{subfigure}
    \caption{Normal 2-car scenario at $t=(\SI{1.26}{s}, \SI{2.88}{s})$}
    \label{fig:two_car_nominal}
    \vspace{-0.2in}
\end{figure}

\begin{figure}
    \centering
   \begin{subfigure}[t]{0.5\columnwidth}
        \centering
        \includegraphics[width=0.98\textwidth, trim={10cm 20cm 20cm 0},clip]{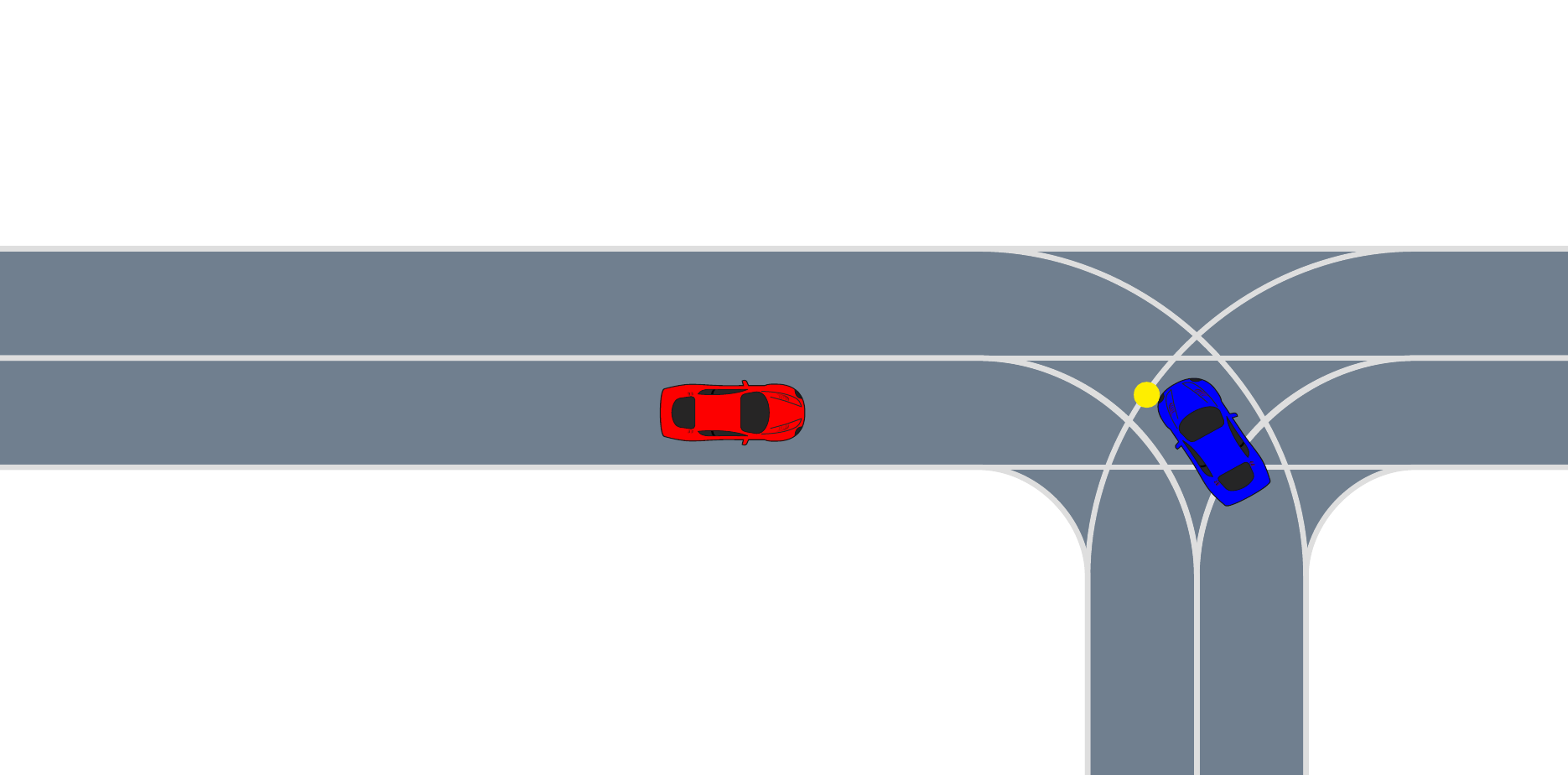}
    \end{subfigure}%
    \begin{subfigure}[t]{0.5\columnwidth}
        \centering
        \includegraphics[width=0.98\textwidth, trim={10cm 20cm 20cm 0},clip]{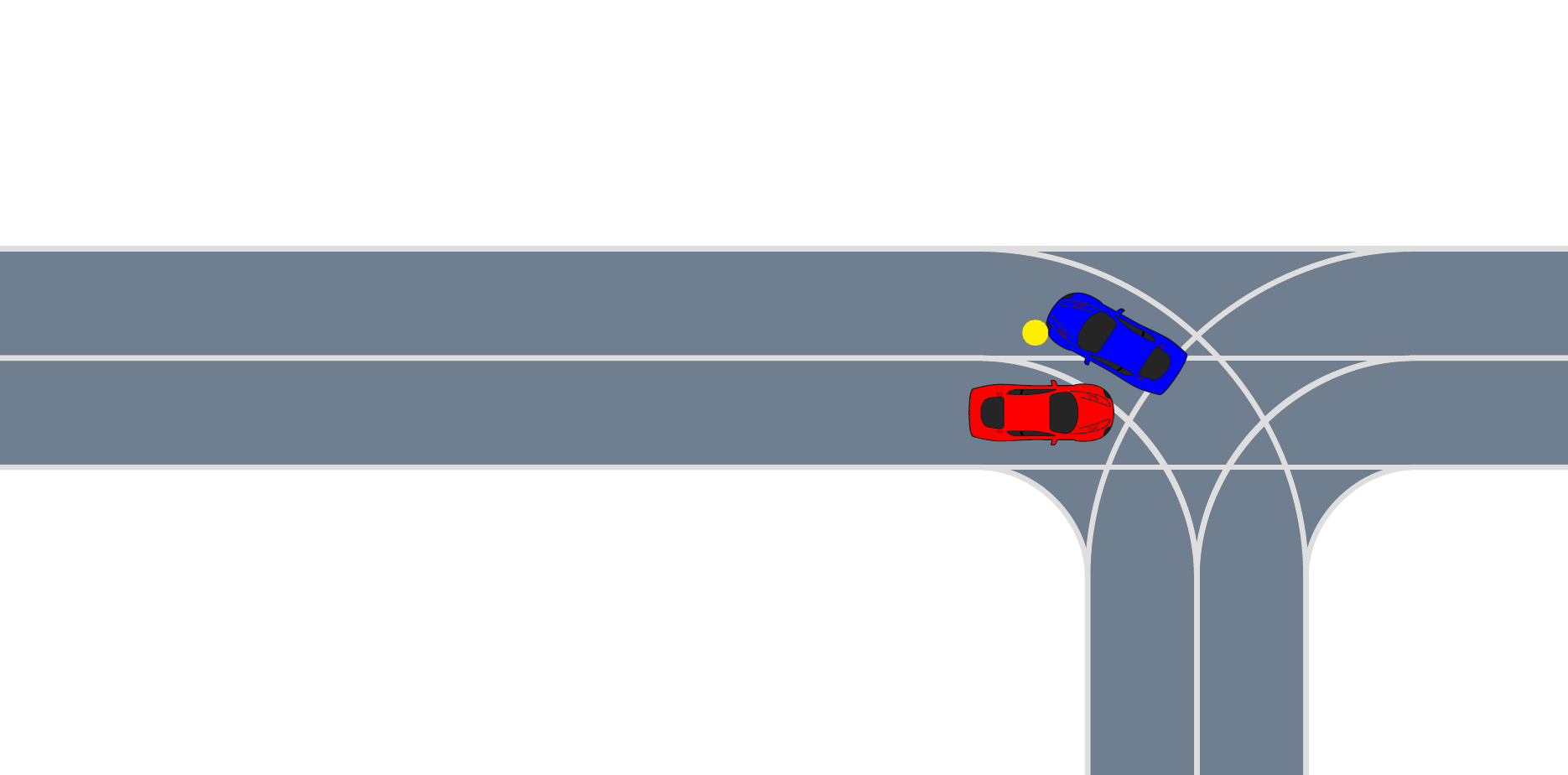}
    \end{subfigure}
    \caption{Collision in 2-car scenario at $t=(\SI{1.26}{s}, \SI{1.80}{s})$}
    \label{fig:two_car_collision}
    \vspace{-0.2in}
\end{figure}

The first scenario is an interaction between the ego vehicle and one adversarial vehicle over a range of initial conditions. \Cref{fig:two_car_nominal} shows one mode of expected behavior in the scenario: the ego vehicle correctly predicts it can cross the intersection before the other driver arrives so it proceeds with the left turn. A sample failure is shown in \cref{fig:two_car_collision}. We can see that the adversary had to accelerate early in the simulation to cause a collision with the ego vehicle, which did not predict that an acceleration would occur. 

\Cref{tab:2car_results} shows the number of failures observed with each approach. Monte Carlo sampling finds the fewest failures with a rate of \num{8e-3} but the failure trajectories have a comparatively large log-likelihood. The uniform importance sampling approach increases the number of failures found by making rare disturbances more likely, but causes the found failures to be extremely unlikely due to these rare disturbances. The cross entropy method finds slightly more failures than the Monte Carlo approach with a larger log likelihood than uniform importance sampling. The DP approach was most successful with a failure rate of \num{1.67e-1} while still retaining a large value of log likelihood. 

\begin{table}
    \centering
    \caption{2-Car Scenario Failure Rates}
    \label{tab:2car_results}
    \begin{tabular}{@{}lll@{}} 
        \toprule
        \textbf{Method} & \textbf{Failure Rate} & \textbf{Log Likelihood}\\
        \midrule
        Monte Carlo & \num{0.008} $\pm$ \num{0.003} & \num{-0.311} $\pm$ \num{0.029} \\
         Uniform Importance Sampling & \num{0.050} $\pm$ \num{0.007} & \num{-5.291} $\pm$ \num{0.065} \\
        Cross Entropy Method & \num{0.009} $\pm$ \num{0.003} & \num{-0.452} $\pm$ \num{0.038} \\
        Dynamic Programming (ours) & \num{0.167} $\pm$ \num{0.012} & \num{-0.656} $\pm$ \num{0.021} \\
        \bottomrule
    \end{tabular}
\end{table}

\subsection{5 Vehicle Interaction}
\begin{figure*}
    \centering
   \begin{subfigure}[t]{0.33\textwidth}
        \centering
        \includegraphics[width=0.8\textwidth,trim={10cm 15cm 15cm 0cm},clip]{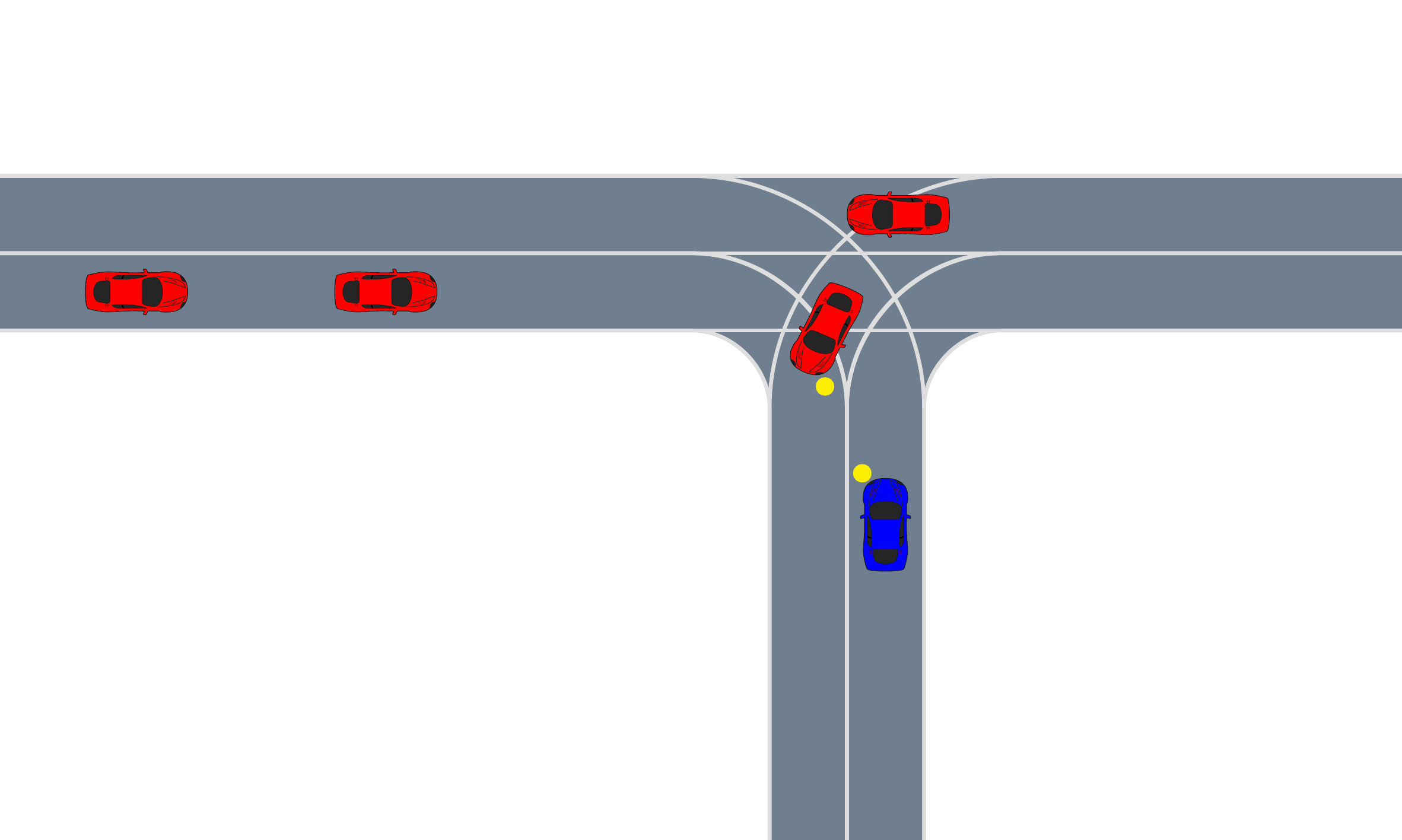}
    \end{subfigure}%
    \begin{subfigure}[t]{0.33\textwidth}
        \centering
        \includegraphics[width=0.8\textwidth,trim={10cm 15cm 15cm 0cm},clip]{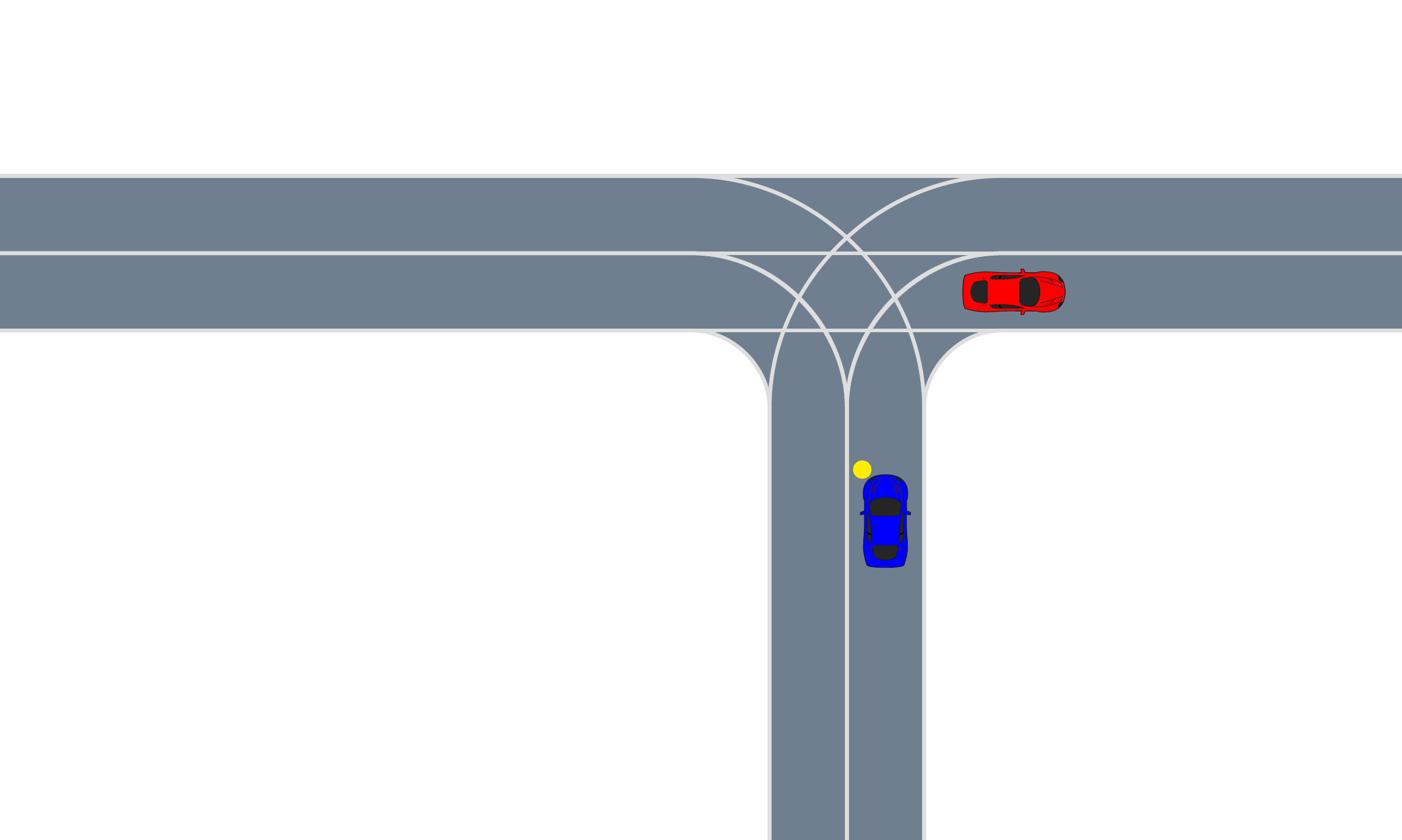}
    \end{subfigure}%
    \begin{subfigure}[t]{0.33\textwidth}
        \centering
        \includegraphics[width=0.8\textwidth,trim={10cm 15cm 15cm 0cm},clip]{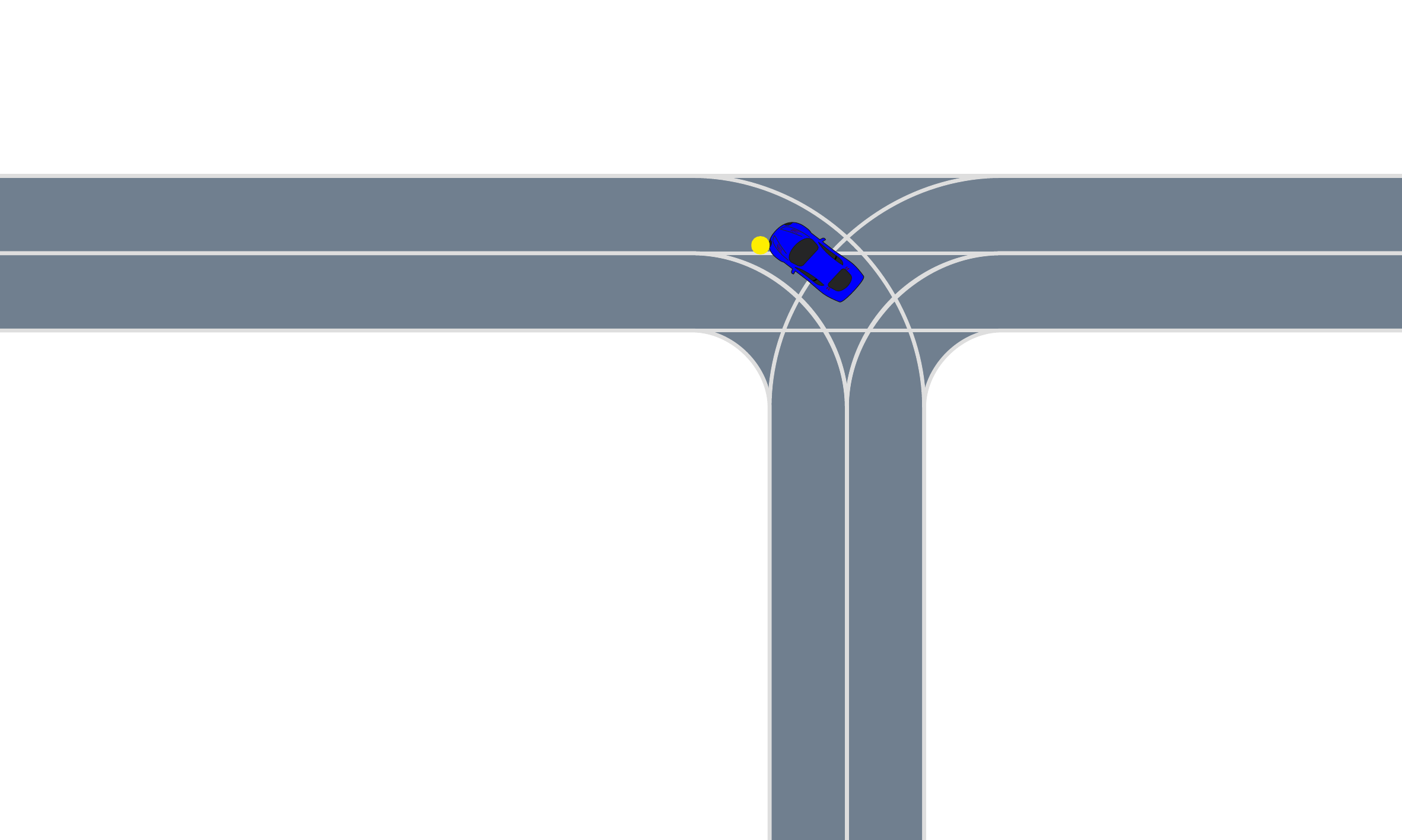}
    \end{subfigure}
    \caption{Normal 5-car scenario at $t=(\SI{0.9}{s}, \SI{4.32}{s}, \SI{7.02}{s})$}
    \label{fig:five_car_nominal}
\end{figure*}

\begin{figure*}
    \centering
   \begin{subfigure}[t]{0.33\textwidth}
        \centering
        \includegraphics[width=0.8\textwidth,trim={10cm 15cm 15cm 0},clip]{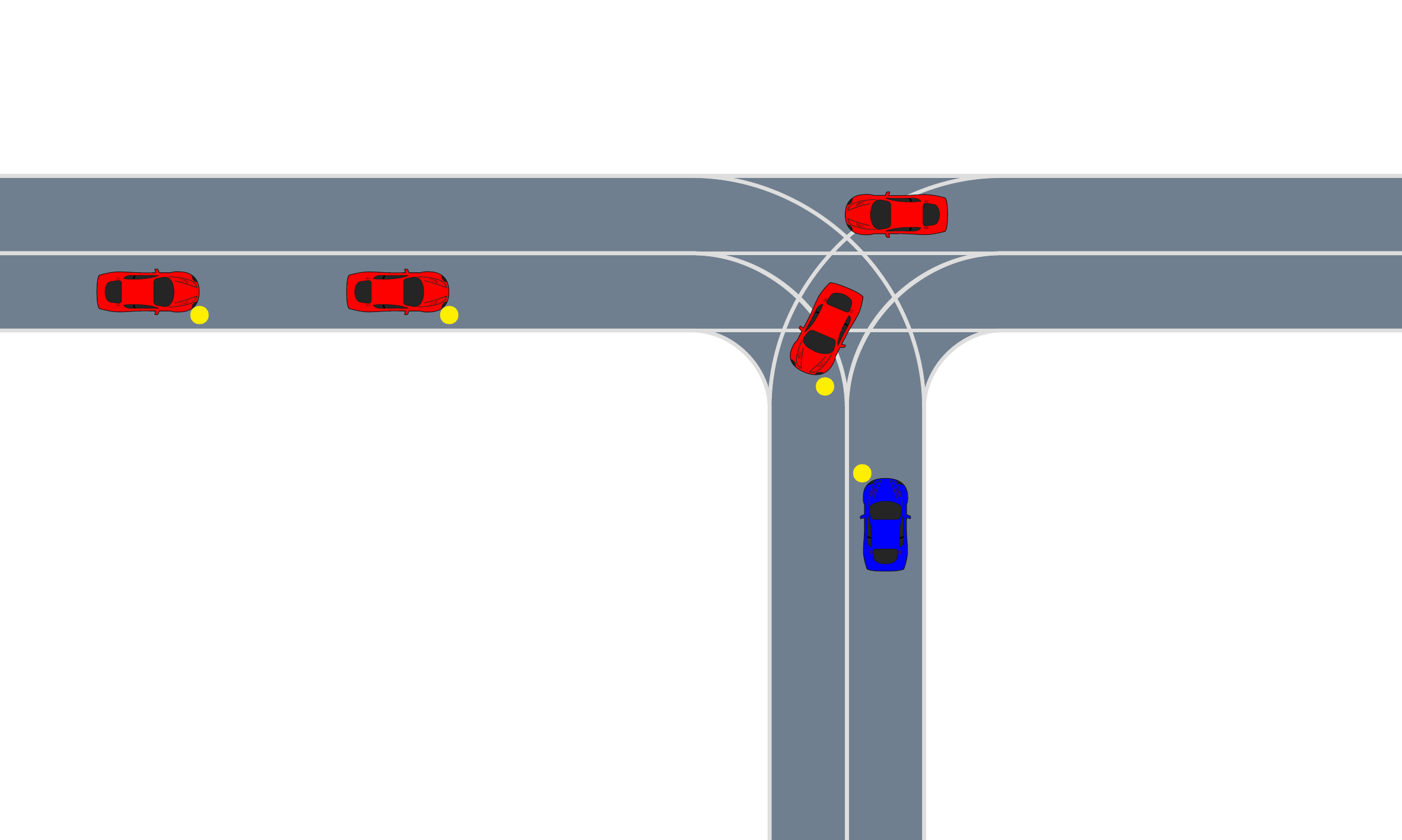}
    \end{subfigure}%
    \begin{subfigure}[t]{0.33\textwidth}
        \centering
        \includegraphics[width=0.8\textwidth,trim={10cm 15cm 15cm 0},clip]{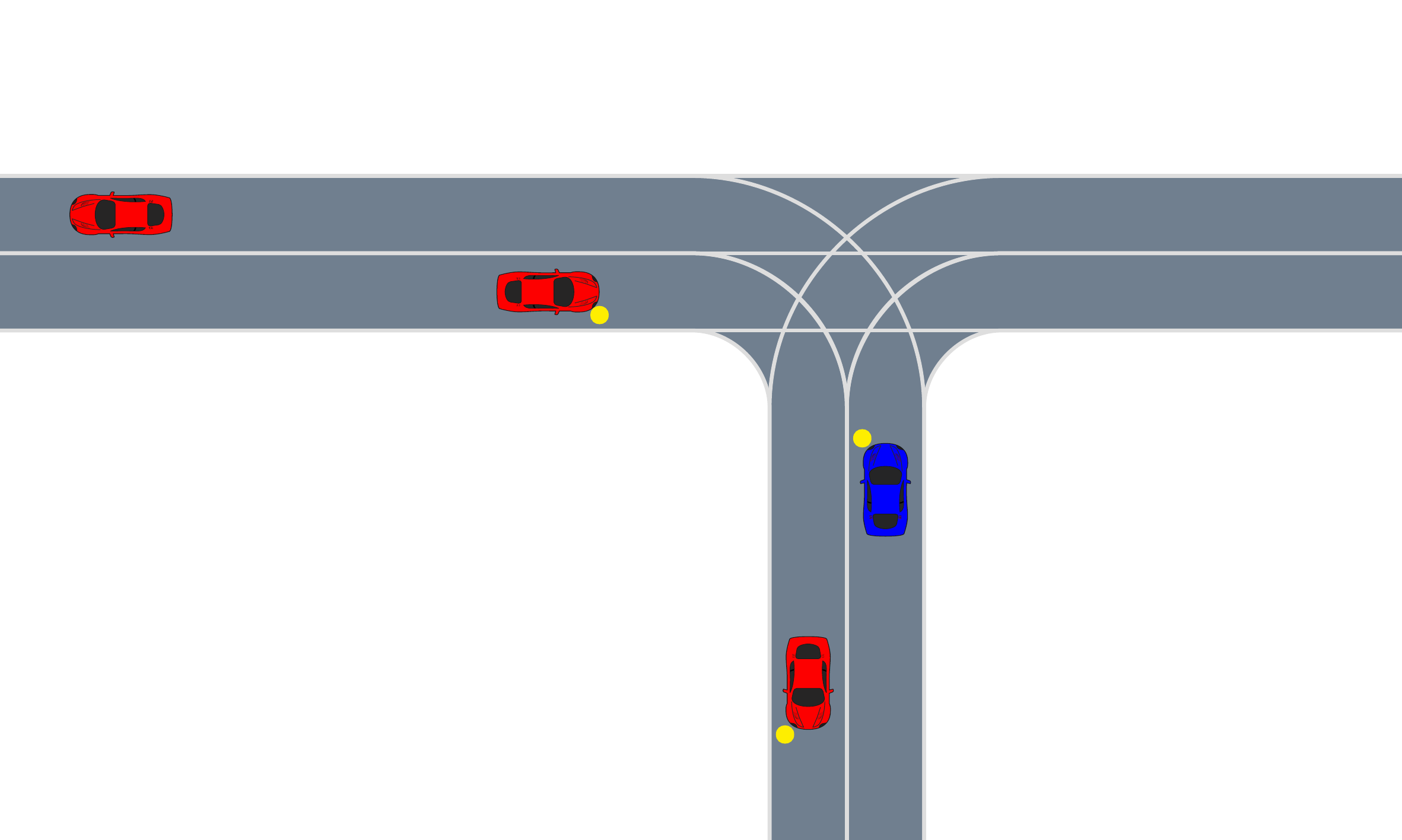}
    \end{subfigure}%
    \begin{subfigure}[t]{0.33\textwidth}
        \centering
        \includegraphics[width=0.8\textwidth,trim={10cm 15cm 15cm 0},clip]{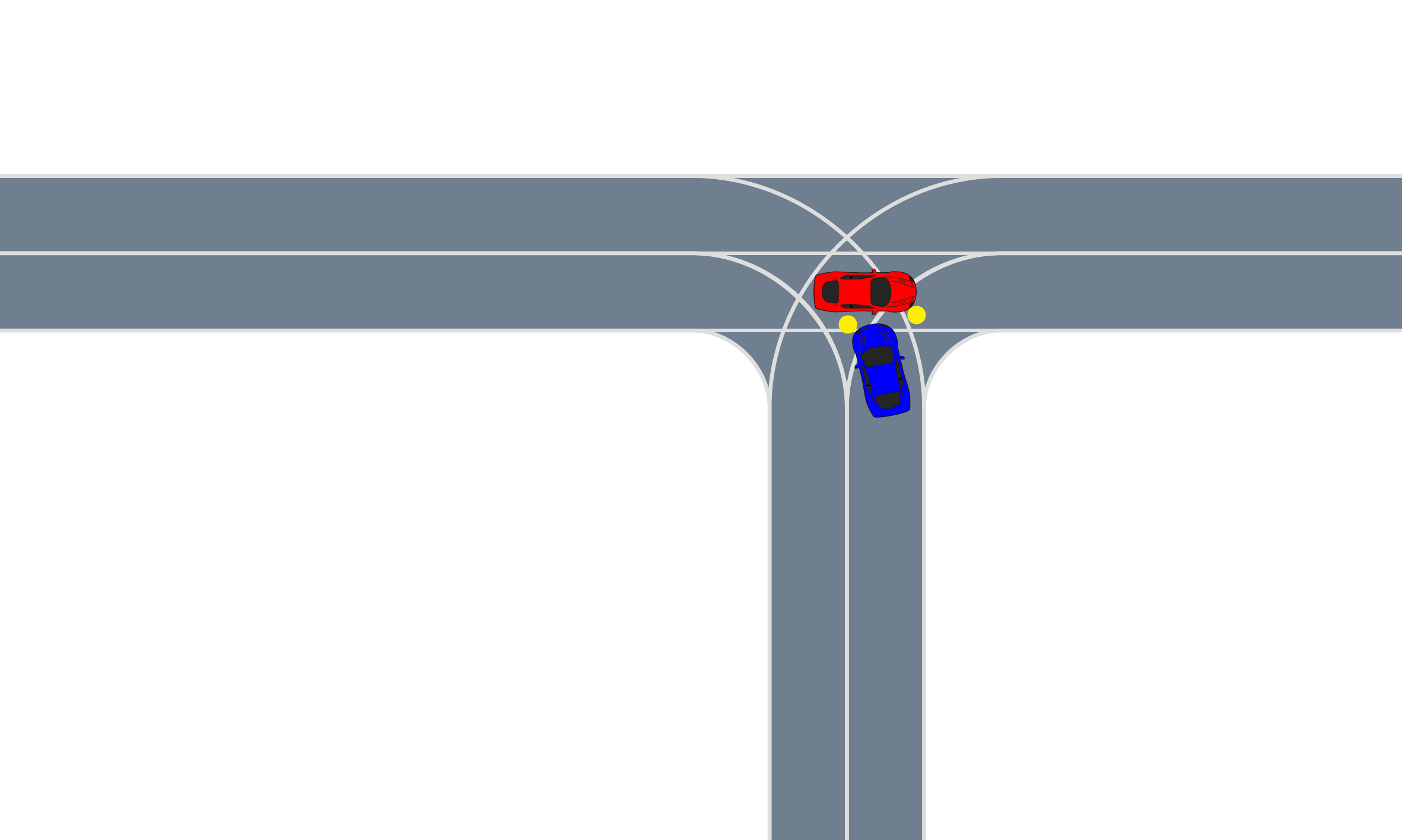}
    \end{subfigure}
    \caption{Collision in 5-car scenario at $t=(\SI{0.9}{s}, \SI{2.34}{s}, \SI{3.42}{s})$}
    \label{fig:five_car_collision}
\end{figure*}

The second scenario involves the interaction of the ego vehicle with four adversarial drivers. A sample of normal behavior for the scenario is shown in \cref{fig:five_car_nominal} where the cars on the left and the trailing car on the right go straight, while the leading car on the right turns onto the vertical road segment. The ego vehicle gives way to all four vehicles and completes the left turn after they have passed. A sample failure is shown in \cref{fig:five_car_collision}. The failure shows that the last car on the left turns it signal on while continuing straight through the intersection, tricking the ego vehicle into initiating the left turn too early.

The driving scenario was broken into four subproblems, one for each adversarial vehicle. The probability of failure was computed for each driving scenario using approximate DP and the solutions were combined using an A2T network trained on rollouts of the full simulator. One challenge for this approach is the exponential scaling of the disturbance space of the full system. If there are 4 subproblems each with \num{7} disturbances then the full problem must consider \num{2401} disturbances per step. To mitigate this problem, we only let one agent act at each timestep, reducing the possible disturbances to \num{28}. This design choice reduces the complexity of possible failure modes, but makes the problem tractable while still finding failures. 

The results are shown in \cref{tab:5car_results}. We first note that the failure rate in the scenario is lower than the previous scenario as indicated by the failure rate of the Monte Carlo approach (\num{1.1e-3}). The uniform importance sampling approach improves failure rate significantly but finds failures with very low likelihood due to the increased number of rare disturbances. The cross entropy method has twice the failure rate as the Monte Carlo approach with a similar log-likelihood. Our approach (DP combined with A2T) has a much larger failure rate (\num{2.22e-1}) while finding relatively likely failures, demonstrating that scene decomposition combined with A2T is an effective strategy for finding failures of an autonomous vehicle in a complex driving scenario. 

\begin{table}
    \centering
    \caption{5-Car Scenario Failure Rates}
    \label{tab:5car_results}
    \begin{tabular}{@{}lll@{}} 
        \toprule
        \textbf{Method} & \textbf{Failure Rate} & \textbf{Log Likelihood}\\
        \midrule
        Monte Carlo & \num{0.0011} $\pm$ \num{0.0003} & \num{-0.437} $\pm$ \num{0.069} \\
         Uniform Importance Sampling & \num{0.033} $\pm$ \num{0.002} & \num{-7.374} $\pm$ \num{0.079} \\
        Cross Entropy Method & \num{0.0027} $\pm$ \num{0.0005} & \num{-0.436} $\pm$ \num{0.041} \\
        A2T + DP (ours) & \num{0.222} $\pm$ \num{0.004} & \num{-0.947} $\pm$ \num{0.145} \\
        \bottomrule
    \end{tabular}
\end{table}

\section{CONCLUSIONS}
\label{sec:conclusions}
In this work, we have made progress toward the goal of automated testing of autonomous vehicles. We introduced a safety validation formulation that uses approximate dynamic programming to estimate the distribution over failures and create sequences of disturbances that cause an autonomous system to fail. The problem of scalability was addressed by decomposing the driving scenario into pairwise interactions between the ego vehicles and other agents on the road. These subproblems were solved and recombined to estimate the probability of failure of the full system. To correct for errors in this estimate, we trained an A2T network with Monte Carlo policy evaluation to weight each subproblem based on the state. We observed \num{1} to \num{2} orders of magnitude increase in the number of failures found compared to importance sampling baselines in a two-vehicle driving scenario and a more complex five-vehicle driving scenario, demonstrating the benefit of this approach. Future work will use the calculated policy to obtain a low-variance estimate of the probability of failure, test performance on more complicated driving scenarios with many agents, and attempt to interpret the attention weights parameters to understand the cause of failures. 

\section*{Acknowledgment}
The authors gratefully acknowledge the financial support from the Stanford Center for AI Safety.  We also thank the NASA AOSP SWS Project.

\renewcommand*{\bibfont}{\footnotesize}
\printbibliography
\end{document}

%% file: A2T_Network.tex
\begin{tikzpicture}
    \tikzstyle{every node}=[font=\small, align=center]
    \tikzset{
        n/.style={draw, rounded corners, minimum height=0.5cm, minimum width = 1.5cm},
        n2/.style={n, minimum width=2.5cm}
        }

    \node (state) [n, rotate=90] {\small State $s$};

    \node (base) [n2, above right of=state, xshift=1.5cm] {Base network};

    \node (solutionsback1) [n2, right of=state, xshift=1.3cm, yshift=-0.1cm] {};
    \node (solutionsback2) [n2, fill=white, right of=state, xshift=1.25cm, yshift=-0.05cm] {};
    \node (solutions) [n2, fill=white, right of=state, xshift=1.2cm] {Solutions};

    \node (attn) [n2, below right of=state, xshift=1.5cm] {Attention network};

    \node (values) [n, below right of=base, xshift=1.8cm, yshift = 0.4cm] {values};
    
     \node (weights) [n, above right of=attn, xshift=1.8cm, yshift = -0.4cm] {weights};
     
     \node (pfail) [n, minimum width = 0.5cm, right of=state, xshift=5.5cm] {$v(s)$};

    \draw[-latex] (state.south) -- (base.west);
    \draw[-latex] (state.south) -- (solutions.west);
    \draw[-latex] (state.south) -- (attn.west);
    
    \draw[-latex] (base.east) -- (values.west);
    \draw[-latex] (solutions.east) -- (values.west);
    \draw[-latex] (attn.east) -- (weights.west);
    
    \draw[-latex] (weights.east) -- (pfail.west);
    \draw[-latex] (values.east) -- (pfail.west);

    \draw [dashed, -latex] (weights.south) to [out=-1500,in=-60] (attn.south);
    \draw [dashed, -latex] (pfail.south) to [out=-150,in=-60] (weights.south);
    \draw [dashed, -latex] (values.north) to [out=150,in=60] (base.north);
    \draw [dashed, -latex] (pfail.north) to [out=150,in=60] (values.north);
\end{tikzpicture}